\documentclass[letterpaper]{article} %
\usepackage{aaai2026}  %
\usepackage{times}  %
\usepackage{helvet}  %
\usepackage{courier}  %
\usepackage[hyphens]{url}  %
\usepackage{graphicx} %
\usepackage{natbib}  %
\usepackage{caption} %
\usepackage{algorithm}
\usepackage{algorithmic}

\usepackage{booktabs}       %
\usepackage{amsfonts}       %
\usepackage{microtype}      %
\usepackage{adjustbox}    %
\usepackage{multirow}     %
\usepackage{arydshln}
\usepackage{amsmath}
\usepackage{colortbl}

\usepackage{newfloat}
\usepackage{listings}
\DeclareCaptionStyle{ruled}{labelfont=normalfont,labelsep=colon,strut=off} %
\floatstyle{ruled}
\newfloat{listing}{tb}{lst}{}
\floatname{listing}{Listing}
\title{First-Order Error Matters: Accurate Compensation\\ for Quantized Large Language Models}

\author {
    Xingyu Zheng\textsuperscript{\rm 1, \rm 2}\equalcontrib, 
    Haotong Qin\textsuperscript{\rm 3}\equalcontrib, 
    Yuye Li\textsuperscript{\rm 4}, 
    Haoran Chu\textsuperscript{\rm 2}, 
    Jiakai Wang\textsuperscript{\rm 5},\\
    Jinyang Guo\textsuperscript{\rm 1, \rm 6}, 
    Michele Magno\textsuperscript{\rm 3}, 
    Xianglong Liu\textsuperscript{\rm 1, \rm 2}\thanks{Corresponding author.}
}
\affiliations {
    \textsuperscript{\rm 1}State Key Laboratory of Complex \& Critical Software Environment, Beihang University \\
    \textsuperscript{\rm 2}School of Computer Science and Engineering, Beihang University \\
    \textsuperscript{\rm 3}ETH Zurich \\
    \textsuperscript{\rm 4}Xidian University \\
    \textsuperscript{\rm 5}Zhongguancun Laboratory\\
    \textsuperscript{\rm 6}School of Artificial Intelligence, Beihang University\\
    \{zhengxingyu,23371505chr,jinyangguo,xlliu\}@buaa.edu.cn, \\
    \{haotong.qin,michele.magno\}@pbl.ee.ethz.ch, liyueye541@gmail.com, wangjk@zgclab.edu.cn
}

\usepackage{bibentry}

\begin{document}

\maketitle

\begin{abstract}
Post-training quantization (PTQ) offers an efficient approach to compressing large language models (LLMs), significantly reducing memory access and computational costs. Existing compensation-based weight calibration methods often rely on a second-order Taylor expansion to model quantization error, under the assumption that the first-order term is negligible in well-trained full-precision models. However, we reveal that the progressive compensation process introduces accumulated first-order deviations between latent weights and their full-precision counterparts, making this assumption fundamentally flawed. To address this, we propose \textbf{FOEM}, a novel PTQ method that explicitly incorporates first-order gradient terms to improve quantization error compensation. 
FOEM approximates gradients by performing a first-order Taylor expansion around the pre-quantization weights. This yields an approximation based on the difference between latent and full-precision weights as well as the Hessian matrix. When substituted into the theoretical solution, the formulation eliminates the need to explicitly compute the Hessian, thereby avoiding the high computational cost and limited generalization of backpropagation-based gradient methods. This design introduces only minimal additional computational overhead.
Extensive experiments across a wide range of models and benchmarks demonstrate that FOEM consistently outperforms the classical GPTQ method. In 3-bit weight-only quantization, FOEM reduces the perplexity of Llama3-8B by 17.3\% and increases the 5-shot MMLU accuracy from 53.8\% achieved by GPTAQ to 56.1\%. Moreover, FOEM can be seamlessly combined with advanced techniques such as SpinQuant, delivering additional gains under the challenging W4A4KV4 setting and further narrowing the performance gap with full-precision baselines, surpassing existing state-of-the-art methods.
  
\end{abstract}

\begin{links}
    \link{Code}{https://github.com/Xingyu-Zheng/FOEM}
\end{links}

\section{Introduction}

Large language models (LLMs), such as Llama~\citep{touvron2023Llama}, have shown remarkable performance and wide-ranging applicability in areas including language understanding~\citep{radford2018improving}, dialogue systems~\citep{chen2017survey}, code generation~\citep{li2022competition}, protein prediction and design~\citep{kuhlman2019advances}, and embodied intelligence~\citep{gupta2021embodied}. As model size and the scale of pretraining data increase, their capabilities continue to improve. However, the substantial number of parameters and high computational requirements impose significant memory and processing burdens. These demands create practical limitations for real-world deployment, especially in resource-constrained environments.

Quantization is a classical model compression technique~\citep{huang2024empirical,gong2024survey,zheng2025empirical,qin2024accurate}. It reduces memory usage and speeds up computation by converting high-bit floating-point parameters and activations into low-bit fixed-point formats, without changing the model architecture. Among various quantization methods, post-training quantization (PTQ)~\citep{lin2024awq,feng2025quantized,lv2024ptq4sam,wang2025SLMQuant} is known for its efficiency. It does not require gradient-based fine-tuning and can maintain nearly lossless performance at higher bit-widths. Compared to quantization-aware training (QAT)~\cite{gholami2022survey,zheng2024bidm,zheng2024binarydm,feng2025mpq,feng2025s,feng2025mpq_v2}, which involves additional training, PTQ is generally more practical for large language models. GPTQ~\citep{frantar2022gptq} is a representative PTQ method for weight-only quantization in large models. It estimates quantization loss using a Taylor expansion, uses second-order information from the Hessian matrix to perform column-wise quantization, and compensates for errors in later columns based on earlier quantization steps. This approach often performs better than simpler methods such as round-to-nearest (RTN).

However, we identify a potential source of error in existing LLM PTQ methods that rely on error reconstruction and compensation. These methods typically assume that the full-precision model has already been fully optimized. Based on this assumption, they omit the first-order term in the loss modeling process. In addition, they use practical approximations to handle the second- and higher-order terms. These simplifications can lead to the accumulation of errors. As a result, the latent weights in later columns, which are calibrated after the earlier ones, may exhibit significant gradients during quantization. If these gradients are ignored, the loss modeling becomes less accurate, which can lead to suboptimal compensation and a drop in overall quantization performance.

In this work, we propose an enhanced method called \textbf{FOEM}, which compensates for output error by incorporating first-order gradient information. Instead of computing gradients through backpropagation, FOEM approximates them using the product of the Hessian matrix and the difference between the compensated latent weights and the original full-precision weights. When this approximation is substituted into the theoretical solution, the additional computation of the Hessian for the compensation term can be eliminated. This effectively avoids the high cost of real-time gradient calculation, making it feasible to integrate gradient-based correction into the PTQ process.

Extensive experiments on the Llama family demonstrate that FOEM outperforms the classic GPTQ~\citep{frantar2022gptq} method at any bit-width. For example, under 3-bit weight-only quantization, it reduces perplexity loss by up to 17.3\%. Moreover, FOEM can be effectively combined with state-of-the-art PTQ techniques such as SpinQuant~\citep{liu2024spinquant}, further advancing the accuracy of LLM quantization. Under the challenging W4A4KV4 setting, where weights, activations, and KV cache are all quantized to 4 bits, our method further reduces the perplexity of Llama3-8B on WikiText2~\cite{merity2016pointer} by 0.20. These results highlight the potential of our approach for enabling more efficient and broadly applicable deployment of large language models.

\section{Related Work}
\label{related}

\paragraph{Post-training Quantization.}
Quantization not only reduces memory consumption by mapping full-precision weights to low-bit fixed-point formats such as int8 or int4, but also enables dynamic quantization of activations into low-bit representations. This facilitates efficient operations, including multiplication of low-bit matrixes. To alleviate accuracy degradation caused by the transition from full-precision to low-bit formats, reconstruction-based methods such as AdaRound~\citep{nagel2020up}, BRECQ~\citep{li2021brecq}, and QDrop~\citep{wei2022qdrop} have been developed. These techniques measure and minimize quantization errors within layers or blocks, demonstrating strong performance on architectures such as ResNet~\citep{he2016deep}. However, because of the substantial computational cost incurred during calibration, these approaches are challenging to apply directly to LLMs, which typically contain billions of parameters.

\paragraph{PTQ Methods for LLMs.}
Numerous PTQ strategies have been proposed to address the outlier characteristics commonly observed in LLM. Some methods preserve outliers by maintaining them in higher-bit precision formats, such as LLM.int8()~\citep{dettmers2022gpt3} and AWQ~\citep{lin2023awq}. Others employ smoothing-based scaling techniques (e.g., SmoothQuant~\citep{xiao2023smoothquant}, OmniQuant~\citep{shao2023omniquant}), rotation-based transformations (e.g., QuaRot~\citep{ashkboos2024quarot}, SpinQuant~\citep{liu2024spinquant}), or channel rearrangement methods (e.g., RPTQ~\citep{yuan2023rptq}). These approaches focus on adjusting the distributional properties of weights and activations and have demonstrated promising results in the quantization of activations.
In contrast to these transformation-based methods, which typically apply scaling or clipping prior to quantization, techniques like GPTQ~\citep{frantar2022gptq} explicitly model the quantization loss and directly modify the fp latent weights during calibration. This loss-aware strategy can be effectively combined with other advanced quantization techniques, such as SpinQuant~\citep{liu2024spinquant}, and has recently led to significant improvements exemplified by methods like GPTAQ~\citep{li2025gptqv2}.

\section{Method}

\begin{figure}[t]
    \centering
    \includegraphics[width=1.\linewidth]{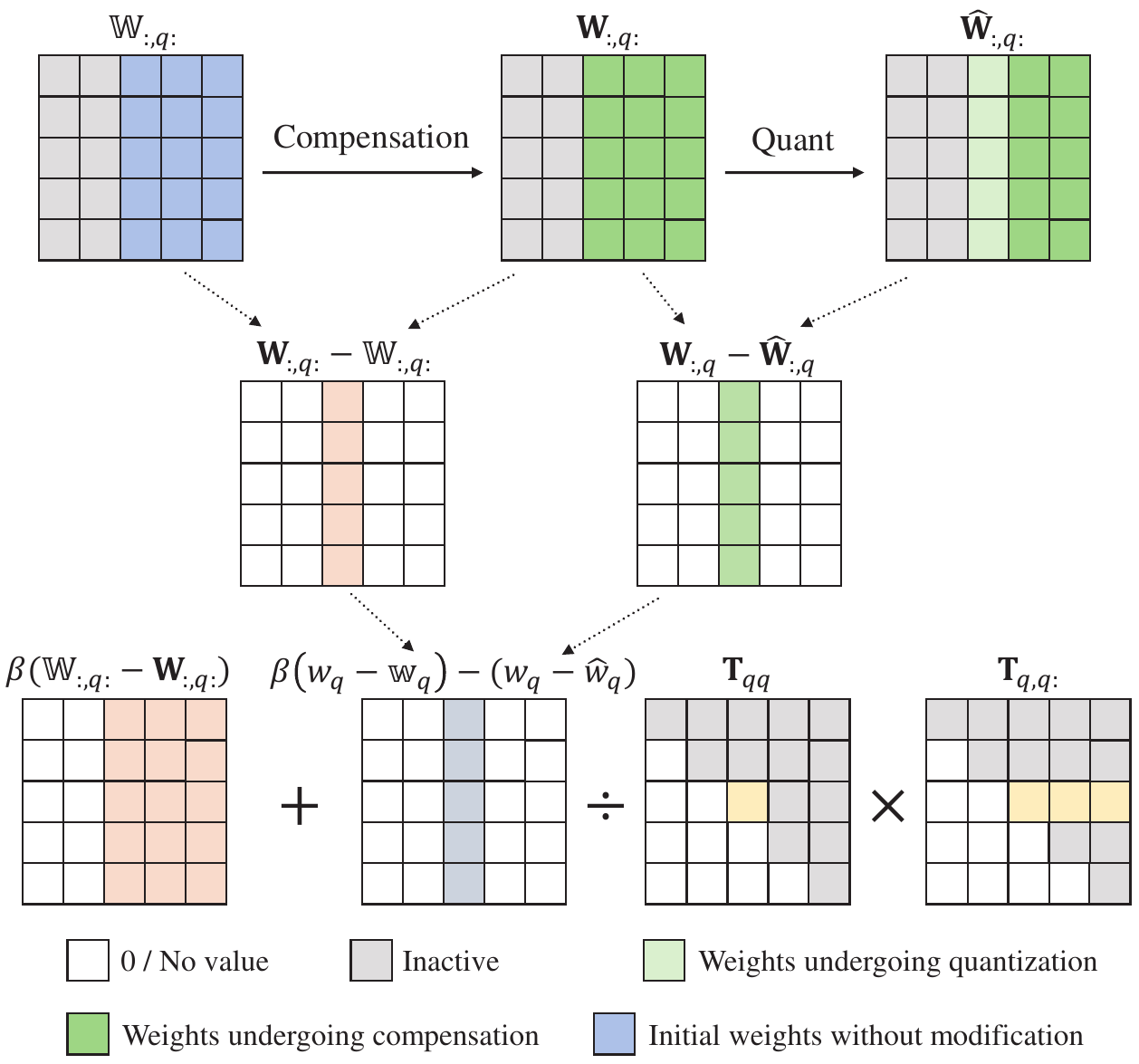}
    \caption{The computation pipeline of our proposed FOEM for the optimal compensation term with gradient consideration.} 
    \label{fig:value}
\end{figure}

\subsection{Preliminaries}

PTQ methods compensated for errors on LLM trace their origins to OBD \citep{lecun1989optimal}, a pruning technique originally developed for small-scale models. Given a layer with original weights $\mathbb{W}$ and input $\mathbf{X}$, and assuming the pruned weights are $\mathbf{W} = \mathbb{W} + \delta \boldsymbol{w}$, the resulting output error $\delta E$ can be approximated using a Taylor series expansion:
\begin{equation}
\delta E = \left( \frac{\partial E}{\partial \boldsymbol w} \right) \delta \boldsymbol w^\top + \frac{1}{2}\delta \boldsymbol w \mathbf H \delta \boldsymbol w^\top + O(\|\delta \boldsymbol w\|^3).
\label{eq:loss}
\end{equation}
OBD neglects higher-order terms and, under the assumption that the model has been well-optimized, treats the first-order term as negligible. In addition, it assumes independence among parameters, considering only the diagonal elements of the Hessian matrix during error estimation.

OBS\citep{hassibi1993optimal} later challenged the independence assumption made in OBD and proposed using the full Hessian matrix to more accurately estimate the error introduced by pruning. Consider the case where the $q$-th parameter is pruned, and the remaining parameters are adjusted to minimize the total loss. The optimization objective can be formulated as:
\begin{equation}
\min_{q} \left\{ \min_{\delta \boldsymbol w} \left( \frac{1}{2}\delta \boldsymbol w \mathbf H \delta \boldsymbol w^\top \mid \boldsymbol e_q \delta \boldsymbol w^\top + \boldsymbol w_q = 0 \right)\right\},
\end{equation}
where $q$ denotes the index of the weight element to be pruned, and $\boldsymbol{e}_q$ is a unit vector with a 1 at the $q$-th position and zeros elsewhere. Applying the method of Lagrange multipliers, the Lagrangian is defined as:
\begin{equation}
\mathcal{L} = \frac{1}{2} \delta \boldsymbol w \mathbf H\delta \boldsymbol w^\top + \lambda (\boldsymbol e_q \delta \boldsymbol w^\top + \boldsymbol w_q),
\end{equation}
where $\lambda$ is the Lagrange multiplier. By taking the partial derivatives with respect to both $\delta \boldsymbol{w}$ and $\lambda$, and setting them to zero, the optimal solution for $\delta \boldsymbol{w}$ can be derived as:
\begin{equation}
\delta \boldsymbol w = -\frac{\boldsymbol w_q}{[\mathbf H^{-1}]_{qq}} [\mathbf H^{-1}]_{q,:}.
\end{equation}

OBC~\cite{frantar2022optimal} observed that computing the inverse Hessian matrix after each pruning step in OBS remains prohibitively expensive for large models. To mitigate this, it restricted optimization to individual rows of the weight matrix. By considering only the second-order term in the Taylor expansion, the objective is reformulated as a sum of output losses across rows:
\begin{equation}
\sum_{i=1}^{d_{\text{raw}}}\left\| \mathbf W_{i,:} \mathbf X - \widehat{\mathbf W}_{i,:} \mathbf X \right\|_2^2.
\end{equation}

It was shown that the Hessian matrix corresponding to each weight row takes the form $\mathbf{H} = 2\mathbf{X} \mathbf{X}^\top$. When a column is pruned from a given weight row, the inverse Hessian for the remaining weights can be efficiently updated through an iterative procedure:
\begin{equation}
\mathbf H_{-p}^{-1} = \left(\mathbf H^{-1} - \frac{1}{[\mathbf H^{-1}]_{qq}} \mathbf H_{:,p}^{-1} \mathbf H_{p,:}^{-1}\right)_{-p}
\label{eq:obc}.
\end{equation}

This pruning-based formulation was further extended to quantization, allowing for the derivation of weight updates for the remaining columns after quantizing a specific weight element:
\begin{equation}
\delta \boldsymbol w = -\frac{\boldsymbol w_q - \hat{\boldsymbol w}_q}{[\mathbf H^{-1}]_{qq}}[\mathbf H^{-1}]_{q,:}.
\end{equation}

GPTQ~\cite{frantar2022gptq} further addressed the computational inefficiencies of OBC when applied to LLMs. It was observed that the order in which the weights are quantized has a minimal effect on the final model performance, allowing it to omit both loss evaluation and sorting operations during weight selection. To improve computational efficiency, GPTQ introduced lazy updates and leveraged Cholesky decomposition.
Specifically, the initial inverse Hessian $\mathbf{H}^{-1}$ is factorized as $\mathbf{H}^{-1} = \mathbf{L} \mathbf{L}^\top$, and the upper triangular matrix $\mathbf{T} = \mathbf{L}^\top$ is retained for subsequent use. This approach avoids iterative updates of $\mathbf{H}^{-1}$ during column-wise calibration and compensation, and leads to the following weight update formula during iterative quantization:
\begin{equation}
\delta \boldsymbol w = -\frac{\boldsymbol w_q - \hat{\boldsymbol w}_q}{\mathbf T_{qq}}\mathbf T_{q,q:}.
\label{eq:gptq}
\end{equation}

\subsection{Analysis of the Neglected First-Order Term}

\begin{algorithm}[t]
   \caption{{\bfseries FOEM} – First-Order Enhanced Method Based on Compensation for quantizing one layer}
   \label{alg:example}
\begin{algorithmic}
   \STATE {\bfseries Input:} FP weight $\mathbf W$, calibration input $\mathbf X$, and Block size $B$
   \STATE {\bfseries Output:} Quantized weight $\boldsymbol Q$
   \STATE $\mathbf H \leftarrow \mathbf X\mathbf X^\top$
   \STATE $\mathbf L=Inverse\_Cholesky(\mathbf H+\lambda_1\boldsymbol I)$
  \STATE {$ \mathbb{W} \leftarrow \mathbf W$}
  \STATE $\boldsymbol Q\leftarrow\boldsymbol{0}_{m\times n}, \boldsymbol E\leftarrow\boldsymbol{0}_{m\times B}$
   \FOR{$i=0, B, 2B, \dots$}
   \FOR{$j=i, i+1, \dots, i+B-1$}
   \STATE $\boldsymbol Q_{:, j}\leftarrow \text{quant}(\mathbf W_{:, j})$
   \STATE $\boldsymbol E_{:, j-i}\leftarrow ((\mathbf W_{:, j} - \boldsymbol Q_{:, j})-\beta(\mathbf W_{:, j} - \mathbb{W}_{:, j}))/\mathbf L_{jj}$
   \STATE $\begin{aligned}
      \mathbf W_{:, j:(i+B)} \leftarrow{} & \mathbf W_{:, j:(i+B)}-\boldsymbol E_{:,j-i}\mathbf L^\top_{j,j:(i+B)} \\
       &-\beta(\mathbf W_{:, j} - \mathbb{W}_{:, j})
   \end{aligned}$
   \ENDFOR
   \STATE $\begin{aligned}
      \mathbf W_{:,(i+B):} \leftarrow{} & \mathbf W_{:,(i+B):}-\boldsymbol E\cdot\mathbf L^\top_{i:(i+B),(i+B):}
   \end{aligned}$
   \ENDFOR
\end{algorithmic}
\end{algorithm}

\begin{table*}[th!]
\centering
\setlength{\tabcolsep}{1.20mm}
\begin{tabular}{ll||c|cc|ccccccc|c}
\hline
\textbf{Model} & \textbf{Method} & \textbf{\#W} & \textbf{Wiki2$\downarrow$} & \textbf{c4$\downarrow$} & \textbf{PiQA} & \textbf{Arc E} & \textbf{Arc C} & \textbf{HellaS} & \textbf{WinoG} & \textbf{BoolQ} & \textbf{Avg$\uparrow$} & \textbf{MMLU$\uparrow$} \\
\hline
\multirow{9}{*}{\centering Llama2-7B} 
& FP16 & 16 & 5.48 & 6.90 & 77.8 & 76.3 & 42.9 & 57.2 & 69.4 & 77.7 & 66.9 & 45.8 \\
\cdashline{2-13}
& RTN & 4 & 5.72 & 7.20 & 77.3 & 75.8 & 42.8 & 56.4 & 68.6 & 77.2 & 66.4 & 44.6 \\
& GPTQ & 4 & 5.61 &7.06 &78.0 &75.9 &42.8 &56.5 &69.6 &78.1 &66.8 &45.0 \\
& GPTAQ & 4 & 5.61 &7.05 &78.1 &75.3 &40.9 &56.4 &69.1 &76.6 &66.1 &44.3
 \\
&\cellcolor{gray!20}FOEM &\cellcolor{gray!20}4 &\cellcolor{gray!20}5.61 &\cellcolor{gray!20}7.05 &\cellcolor{gray!20}77.8 &\cellcolor{gray!20}75.6 &\cellcolor{gray!20}43.3 &\cellcolor{gray!20}56.5 &\cellcolor{gray!20}68.9 &\cellcolor{gray!20}77.7 &\cellcolor{gray!20}66.6 &\cellcolor{gray!20}45.5
 \\
\cdashline{2-13}

& RTN & 3 & 6.92 & 8.66 & 75.5 & 72.5 & 39.2 & 53.8 & 67.5 & 68.0 & 62.7 & 36.1 \\
& GPTQ & 3 & 6.38 &7.85 &76.0 &71.8 &39.8 &54.2 &66.3 &73.1 &63.5 &40.3
\\
& GPTAQ & 3 & 6.41 &7.93 &76.6 &73.3 &40.4 &54.1 &66.5 &73.3 &64.0 &35.1
 \\
\rowcolor{gray!20} \cellcolor{white} & FOEM & 3 & 6.27 &7.81 &77.0 &74.6 &41.0 &54.2 &66.5 &73.4 &64.5 &42.0
 \\
\hline
\multirow{9}{*}{\centering Llama2-13B} 
& FP16 & 16 & 4.89 & 6.41 & 78.9 & 79.3 & 48.1 & 60.1 & 72.3 & 80.6 & 69.9 & 55.2 \\
\cdashline{2-13}
& RTN & 4 & 5.03 & 6.58 & 78.8 & 79.7 & 47.5 & 59.8 & 71.0 & 80.3 & 69.5 & 53.6 \\
& GPTQ & 4 &  5.00 &6.50 &78.9 &78.5 &47.7 &59.8 &72.2 &80.0 &69.5 &54.9
\\
& GPTAQ & 4 & 4.99 &6.50 &78.6 &78.7 &47.4 &59.7 &72.1 &80.5 &69.5 &55.2
 \\
 &\cellcolor{gray!20}FOEM &\cellcolor{gray!20}4 &\cellcolor{gray!20}4.99 &\cellcolor{gray!20}6.50 &\cellcolor{gray!20}79.4 &\cellcolor{gray!20}78.3 &\cellcolor{gray!20}47.8 &\cellcolor{gray!20}59.9 &\cellcolor{gray!20}72.5 &\cellcolor{gray!20}80.1 &\cellcolor{gray!20}69.7 &\cellcolor{gray!20}55.5
 \\
\cdashline{2-13}
& RTN & 3 & 5.61 & 7.21 & 77.6 & 77.2 & 43.7 & 56.5 & 69.4 & 76.0 & 66.7 & 49.0 \\
& GPTQ & 3 &  5.42 &6.94 &78.2 &76.3 &44.4 &58.1 &70.6 &77.7 &67.5 &51.7
\\
& GPTAQ & 3 & 5.42 &6.94 &78.2 &77.2 &43.9 &57.3 &71.0 &78.9 &67.8 &51.5
 \\
\rowcolor{gray!20} \cellcolor{white} & FOEM & 3 & 5.42 &6.93 &78.5 &77.9 &44.6 &58.2 &70.9 &78.5 &68.1 &51.8
  \\
\hline
\end{tabular}
\caption{Comparison of weight-only quantization for Llama-2 models.}
\label{tab:weight_only}
\end{table*}

The aforementioned methods commonly assume that well-trained models have nearly converged to local optima, thereby justifying the omission of first-order terms in the loss approximation. However, we observe that, following the quantization of preceding weight columns, the remaining unquantized weights $\mathbf{W}$ can deviate significantly from their original full-precision values $\mathbb{W}$ due to successive compensations $\delta \boldsymbol{w}$. This accumulated deviation can lead to non-negligible gradients $\boldsymbol{g} = \frac{\partial E}{\partial \boldsymbol{w}}$, making the first-order term in the Taylor expansion a potentially significant contributor to the quantized loss function.

Therefore, we begin by retaining the first-order term in the loss formulation of Eq.~\eqref{eq:loss} and evaluate its impact on the final loss and compensation results. We formulate the pre- and post-quantization loss as:
\begin{equation}
\delta E = \boldsymbol g \delta \boldsymbol w^\top + \frac{1}{2} \delta \boldsymbol w \mathbf H\delta \boldsymbol w^\top.
\end{equation}

When quantizing the $q$-th weight column, the optimization objective becomes minimizing the quantization-induced loss by adjusting the latent weights $\delta \boldsymbol{w}$ of the remaining unquantized columns:
\begin{equation}
\min_{\delta \boldsymbol w} \left( \boldsymbol g \delta \boldsymbol w^\top + \frac{1}{2}\delta \boldsymbol w \mathbf H \delta \boldsymbol w^\top \mid \boldsymbol e_q \delta \boldsymbol w^\top + \boldsymbol w_q - \hat{\boldsymbol w}_q = 0 \right),
\end{equation}
where $\hat{\boldsymbol{w}}_q$ denotes the quantized value of the $q$-th weight column and $\boldsymbol{e}_q$ is a unit vector with 1 at the $q$-th position.

To solve this constrained optimization problem, we apply the method of Lagrange multipliers and define the Lagrangian:
\begin{equation}
\mathcal{L} = \boldsymbol g \delta \boldsymbol w^\top + \frac{1}{2} \delta \boldsymbol w \mathbf H\delta \boldsymbol w^\top + \lambda (\boldsymbol e_q \delta \boldsymbol w^\top + \boldsymbol w_q - \hat{\boldsymbol w}_q).
\end{equation}

Taking derivatives with respect to $\delta \boldsymbol w$ and $\lambda$ yields:
\begin{equation}
\left\{
\begin{aligned}
&\frac{\partial \mathcal{L}}{\partial \delta \boldsymbol w} =\boldsymbol g + \delta \boldsymbol w \mathbf H + \lambda \boldsymbol e_q \\
&\frac{\partial \mathcal{L}}{\partial \lambda} = \boldsymbol e_q \delta \boldsymbol w^\top + \boldsymbol w_q - \hat{\boldsymbol w}_q
\end{aligned}
\right..
\end{equation}

Setting these derivatives to zero gives the optimal $\delta \boldsymbol w$:
\begin{equation}
\delta \boldsymbol w = -\frac{\boldsymbol w_q - \hat{\boldsymbol w}_q-\boldsymbol gH^{-1}e_q^\top}{[\mathbf H^{-1}]_{qq}}[\mathbf H^{-1}]_{q,:} - \boldsymbol g\mathbf H^{-1}.
\end{equation}

By incorporating GPTQ’s Cholesky decomposition, where the inverse Hessian $\mathbf{H}^{-1} = \mathbf{L} \mathbf{L}^\top$ and $\mathbf{T} = \mathbf{L}^\top$ is retained as an upper triangular matrix, the update further simplifies to:
\begin{equation}
\delta \boldsymbol w = -\frac{\boldsymbol w_q - \hat{\boldsymbol w}_q-\boldsymbol gH^{-1}e_q^\top}{\mathbf T_{qq}}\mathbf T_{q,q:} - \boldsymbol g\mathbf H^{-1}.
\label{eq:real}
\end{equation}

Compared to second-order-only approaches such as Eq.~\eqref{eq:gptq}, our formulation introduces additional terms involving $g$, which account for the compensation required due to previously quantized columns.

However, directly computing this term presents practical challenges: iterative updates of $\mathbf{H}^{-1}$, as described in Eq.~\eqref{eq:obc}, are computationally prohibitive for large language models, and obtaining the gradient $\boldsymbol{g}$ via backpropagation is similarly infeasible due to the high memory and compute cost.

\subsection{Practical First-Order Error Compensation}

To overcome these computational challenges, we approximate the gradient $\boldsymbol{g}$ using the difference between the full-precision weights and their compensated versions prior to quantization.

\paragraph{Gradient Approximation.}
While computing exact gradients via backpropagation after each compensation step for the remaining quantized weights $\mathbf{W}$ is computationally infeasible, we observe that the first-order gradient is closely related to the deviation between $\mathbf{W}$ and the original full-precision weights $\mathbb{W}$. By performing a Taylor expansion of the loss around $\mathbf{W}$, we can approximate the gradient on $\mathbf{W}$ as:
\begin{equation}
\boldsymbol g(\mathbf W) \approx \boldsymbol g(\mathbb{W}) + ( \mathbf W - \mathbb{W})\mathbf{H}.
\end{equation}

Since the initial weights $\mathbb{W}$ are assumed to be pretrained to optimality, we have:
\begin{equation}
\boldsymbol g(\mathbb{W}) \approx 0.
\end{equation}

Therefore, we can obtain an approximate estimate of $\boldsymbol{g}(\mathbf{W})$:
\begin{equation}
\boldsymbol g(\mathbf W) \approx ( \mathbf W - \mathbb{W})\mathbf{H}.
\end{equation}

Furthermore, from a practical perspective, large gradients can significantly affect the optimization results and potentially amplify approximation errors. To address this, we introduce an empirical stabilization factor $\beta$, ensuring that the inclusion of the first-order term consistently leads to improvements. Therefore, the gradient used in practice is computed as follows:
\begin{equation}
g = \beta( \mathbf W - \mathbb{W})\mathbf{H},
\label{eq:approx}
\end{equation}
where $\beta$ is typically set to 0.1, consistently leading to stable improvements.

\begin{table*}[th!]
\centering
\setlength{\tabcolsep}{1.20mm}
\begin{tabular}{ll||c|cc|ccccccc|c}
\hline
\textbf{Model} & \textbf{Method} & \textbf{\#W} & \textbf{Wiki2$\downarrow$} & \textbf{c4$\downarrow$} & \textbf{PiQA} & \textbf{Arc E} & \textbf{Arc C} & \textbf{HellaS} & \textbf{WinoG} & \textbf{BoolQ} & \textbf{Avg$\uparrow$} & \textbf{MMLU$\uparrow$} \\
\hline
\multirow{9}{*}{\centering Llama3-8B} 
& FP16 & 16 & 6.13 & 9.61 & 79.5 & 80.1 & 50.1 & 60.1 & 73.3 & 81.0 & 70.7 & 64.9 \\
\cdashline{2-13}
& RTN & 4 & 6.57 & 10.15 & 79.0 & 79.2 & 47.9 & 59.3 & 73.1 & 80.5 & 69.8 & 63.0 \\
& GPTQ & 4 & 6.54 &10.13 &78.2 &78.1 &47.5 &59.2 &73.8 &80.8 &69.6 &63.2\\
& GPTAQ & 4 & 6.61 &10.21 &78.2 &78.5 &48.0 &59.3 &73.3 &81.2 &69.8 &62.9
 \\
&\cellcolor{gray!20}FOEM &\cellcolor{gray!20}4 &\cellcolor{gray!20}6.54 &\cellcolor{gray!20}10.13 &\cellcolor{gray!20}78.9 &\cellcolor{gray!20}78.8 &\cellcolor{gray!20}49.1 &\cellcolor{gray!20}59.1 &\cellcolor{gray!20}73.6 &\cellcolor{gray!20}80.6 &\cellcolor{gray!20}70.0 &\cellcolor{gray!20}63.2
 \\
\cdashline{2-13}
& RTN & 3 & 13.10 & 20.50 & 70.3 & 61.2 & 30.9 & 47.4 & 65.1 & 68.0 & 57.2 & 38.9 \\
& GPTQ & 3 & 9.86 &12.94 &76.0 &71.8 &41.4 &55.8 &69.9 &71.8 &64.4 &55.4 \\
& GPTAQ & 3 & 8.92 &12.82 &74.9 &70.4 &38.3 &55.8 &69.9 &70.6 &63.3 &53.8 \\
\rowcolor{gray!20} \cellcolor{white} & FOEM & 3 & 8.32 &12.37 &76.8 &75.2 &44.6 &56.3 &70.9 &69.1 &65.5 &56.1 \\
\hline

\multirow{9}{*}{\centering Llama3.2-1B} 
& FP16 & 16 & 9.75 & 13.90 & 74.4 & 65.5 & 31.4 & 47.7 & 60.3 & 63.8 & 57.2 & 30.9 \\
\cdashline{2-13}
& RTN & 4 & 11.92 & 17.33 & 72.4 & 60.6 & 30.7 & 45.2 & 57.9 & 57.2 & 54.0 & 28.1 \\
& GPTQ & 4 & 10.69 &15.04 &73.9 &63.5 &30.1 &46.6 &60.6 &60.6 &55.9 &28.5
 \\
& GPTAQ & 4 & 10.62 &15.10 &73.7 &63.7 &31.5 &47.0 &61.4 &61.8 &56.5 &28.6
 \\
&\cellcolor{gray!20}FOEM &\cellcolor{gray!20}4 &\cellcolor{gray!20}10.58 &1\cellcolor{gray!20}5.02 &\cellcolor{gray!20}74.3 &\cellcolor{gray!20}62.8 &\cellcolor{gray!20}31.1 &\cellcolor{gray!20}46.7 &\cellcolor{gray!20}57.9 &\cellcolor{gray!20}64.3 &\cellcolor{gray!20}56.2 &\cellcolor{gray!20}29.0
 \\
\cdashline{2-13}
& RTN & 3 & 56.41 & 74.73 & 63.4 & 43.6 & 22.5 & 32.6 & 50.0 & 53.5 & 44.3 & 25.5 \\
& GPTQ & 3 & 16.23 &20.93 &67.9 &57.0 &29.4 &42.3 &57.8 &55.0 &51.5 &25.6
 \\
& GPTAQ & 3 & 15.40 &20.46 &69.5 &54.9 &28.6 &42.4 &55.2 &56.4 &51.2 &26.1
 \\
\rowcolor{gray!20} \cellcolor{white} & FOEM & 3 & 15.11 &20.48 &69.0 &56.7 &26.8 &42.1 &57.1 &57.3 & 51.5 &26.9
 \\
\hline
\multirow{9}{*}{\centering Llama3.2-3B} 
& FP16 & 16 & 7.81 & 11.30 & 76.4 & 74.3 & 42.2 & 55.3 & 69.4 & 72.8 & 65.1 & 56.5 \\
\cdashline{2-13}
& RTN & 4 & 8.69 & 12.74 & 76.4 & 71.8 & 42.4 & 53.6 & 68.4 & 72.3 & 64.2 & 51.9 \\
& GPTQ & 4 & 9.32 &11.99 &76.4 &73.3 &42.4 &55.0 &68.7 &72.4 &64.7 &54.2
 \\
& GPTAQ & 4 & 8.66 &12.02 &76.9 &73.7 &41.8 &54.9 &68.7 &72.5 &64.7 &53.9
 \\
&\cellcolor{gray!20}FOEM &\cellcolor{gray!20}4 &\cellcolor{gray!20}8.67 &\cellcolor{gray!20}11.98 &\cellcolor{gray!20}76.8 &\cellcolor{gray!20}73.6 &\cellcolor{gray!20}42.7 &\cellcolor{gray!20}54.6 &\cellcolor{gray!20}68.4 &\cellcolor{gray!20}72.4 &\cellcolor{gray!20}64.8 &\cellcolor{gray!20}54.4
 \\
\cdashline{2-13}
& RTN & 3 & 18.43 & 25.38 & 69.1 & 60.2 & 29.9 & 43.3 & 60.9 & 54.7 & 53.0 & 29.8 \\
& GPTQ & 3 & 16.51 &14.90 &73.6 &64.3 &34.7 &50.6 &65.5 &61.6 &58.4 &43.5\\
& GPTAQ & 3 & 14.79 &15.03 &74.3 &65.2 &34.6 &50.9 &65.4 &61.8 &58.7 &44.4
 \\
\rowcolor{gray!20} \cellcolor{white} & FOEM & 3 & 14.16 &14.87 &73.6 &64.4 &34.8 &50.5 &63.5 &67.7 &59.1 &46.7
 \\
\hline
\end{tabular}
\caption{Comparison of weight-only quantization for Llama-3 models.}
\label{tab:weight_only_1}
\end{table*}

\paragraph{Final Optimization Result}
By substituting the approximation from Eq.\eqref{eq:approx} into the theoretical expression in Eq.\eqref{eq:real}, the final compensation term, based on Eq.\eqref{eq:gptq} and the Cholesky decomposition used in the original GPTQ, can be expressed as:
\begin{equation}
\delta \boldsymbol w = -\frac{(w_q-\hat w_q)-\beta(w_q-\mathbb{W}e_q^\top)}{\mathbf T_{qq}}\mathbf T_{q,q:}-\beta(\mathbf W-\mathbb{W}).
\label{eq:ours_final}
\end{equation}

Note that the computationally expensive higher-order terms in Eq.~\eqref{eq:approx} and Eq.~\eqref{eq:real}, including the full Hessian $\mathbf{H}$ and its inverse $\mathbf{H}^{-1}$, cancel out upon substitution. This not only removes the costly backpropagation for gradient computation but also eliminates the need to explicitly compute or invert the Hessian. Consequently, by leveraging the triangular matrices from Cholesky decomposition, only a lightweight weight difference computation is required, resulting in negligible additional overhead.
Compared to the original GPTQ, the effect of our first-order correction is reflected in the additional subtracted terms in Eq.~\eqref{eq:ours_final}. Core GPTQ strategies, such as the lazy update mechanism, remain fully compatible and can be used without modification. These extensions and distinctions are summarized in Algorithm~\ref{alg:example}.

\begin{table*}[th!]
\centering
\setlength{\tabcolsep}{1.3mm}
\begin{tabular}{ll||c|cc|ccccccc|c}
\hline
\textbf{Model} & \textbf{Method} & \textbf{\#W} & \textbf{Wiki2$\downarrow$} & \textbf{c4$\downarrow$} & \textbf{PiQA} & \textbf{Arc E} & \textbf{Arc C} & \textbf{HellaS} & \textbf{WinoG} & \textbf{BoolQ} & \textbf{Avg$\uparrow$} & \textbf{MMLU$\uparrow$} \\
\hline

\multirow{5}{*}{\centering Qwen3-8B} 
& FP16 & 16 & 7.00 & 10.43 & 79.3 & 81.9 & 52.7 & 58.9 & 72.1 & 83.1 & 71.3 & 76.7 \\
\cdashline{2-13}
& RTN & 3 & 11.14 & 14.61 & 75.6 & 74.0 & 43.7 & 53.4 & 65.1 & 84.2 & 66.0 & 67.3 \\
& GPTQ & 3 &8.32 &11.55 &77.5 &78.8 &50.6 &55.7 &68.1 &76.9 &68.0 &67.3 \\
& GPTAQ & 3 & 8.20 &11.54 &78.5 &77.0 &47.3 &55.5 &68.1 &72.5 &66.5 &69.6 \\
\rowcolor{gray!20} \cellcolor{white} & FOEM& 3 & 8.17 &11.52 &77.3 &78.1 &47.1 &55.6 &69.2 &81.7 &68.2 &69.7 \\
\hline
\multirow{5}{*}{\centering Phi-1.5B} 
& FP16 & 16 & 21.84 & 20.42 & 76.7 & 76.2 & 44.9 & 47.9 & 73.1 & 74.6 & 65.6 & 42.8 \\
\cdashline{2-13}
& RTN & 3 & 27.20 & 23.80 & 74.9 & 72.1 & 42.4 & 45.1 & 69.3 & 71.4 & 62.5 & 36.5 \\
& GPTQ & 3 & 23.93 & 21.80 & 75.6 & 73.3 & 41.7 & 46.1 & 69.7 & 73.6 & 63.3 &39.5 \\
& GPTAQ & 3 & 24.10 &21.97 &73.8 &73.4 &41.6 &45.7 &69.9 &73.8 &63.0 &39.9 \\
\rowcolor{gray!20} \cellcolor{white} & FOEM& 3 & 23.89 &21.72 &75.2 &73.4 &43.6 &45.9 &70.8 &73.3 &63.7 & 40.1 \\
\hline
\multirow{5}{*}{\centering  Mistral-7B} 
& FP16 & 16 & 5.26 & 7.55 & 80.4 & 80.6 & 50.9 & 61.2 & 74.0 & 83.7 & 71.8 & 62.4 \\
\cdashline{2-13}
& RTN & 3 & 6.70 & 9.03 & 78.6 & 76.1 & 44.6 & 57.7 & 69.1 & 77.1 & 67.2 & 53.0 \\
& GPTQ & 3 & 6.04 &8.86 &79.4 &78.2 &47.4 &58.9 &72.7 &80.9 &69.6 &55.3
 \\
& GPTAQ & 3 & 6.05 &8.85 &79.8 &78.2 &48.3 &59.1 &72.5 &80.9 &69.8 &54.5
 \\
\rowcolor{gray!20} \cellcolor{white} & FOEM & 3 & 6.03 &8.85 &78.9 &78.3 &49.5 &59.4 &72.4 &81.5 &70.0 &55.5
\\
\hline

\end{tabular}
\caption{Comparison of weight-only quantization for a wide variety of models.}
\label{tab:weight_only_2}
\end{table*}

\begin{table*}[th!]
\centering
\setlength{\tabcolsep}{1.55mm}
\begin{tabular}{ll||cc|ccccccc|c}
\hline
\textbf{Model} & \textbf{Method} & \textbf{Wiki2$\downarrow$} & \textbf{c4$\downarrow$} & \textbf{PiQA} & \textbf{Arc E} & \textbf{Arc C} & \textbf{HellaS} & \textbf{WinoG} & \textbf{BoolQ} & \textbf{Avg$\uparrow$} & \textbf{MMLU$\uparrow$} \\
\hline
\multirow{4}{*}{\centering Llama2-7B} 
& FP16 & 5.48 & 6.90 & 77.8 & 76.3 & 42.9 & 57.2 & 69.4 & 77.7 & 66.9 & 45.8 \\
\cdashline{2-12}
& GPTQ & 6.69 &8.52 &74.8 &71.3 &39.6 &51.9 &64.4 &72.2 &62.4 &35.9
 \\
& GPTAQ & 6.66 &8.38 &74.9 &73.2 &38.7 &52.1 &63.1 &71.9 &62.3 &36.5
 \\
\rowcolor{gray!20} \cellcolor{white} & FOEM & 6.55 &8.29 &74.6 &71.5 &39.5 &52.7 &63.8 &73.2 &62.6 &36.9
 \\
\hline
\multirow{4}{*}{\centering Llama2-13B} 
& FP16 & 4.89 & 6.41 & 78.9 & 79.3 & 48.1 & 60.1 & 72.3 & 80.6 & 69.9 & 55.2 \\
\cdashline{2-12}
& GPTQ & 5.67 &7.43 &77.3 &75.2 &42.7 &56.4 &69.1 &79.5 &66.7 &47.5
 \\
& GPTAQ & 5.68 &7.38 &76.9 &75.8 &43.0 &56.4 &69.9 &77.6 &66.6 &47.5
 \\
\rowcolor{gray!20} \cellcolor{white} & FOEM &5.62 &7.36 &77.7& 75.4 &42.7 &56.2 &69.5 &78.5 &66.7 &47.5
\\
\hline
\multirow{4}{*}{\centering Llama3-8B} 
& FP16 & 6.13 & 9.61 & 79.5 & 80.1 & 50.1 & 60.1 & 73.3 & 81.0 & 70.7 & 64.9 \\
\cdashline{2-12}
& GPTQ & 8.55 &13.24 &73.6 &72.3 &42.0 &54.1 &68.0 &74.6 &64.1 &49.6 \\
& GPTAQ & 8.50 &13.13 &74.5 &73.2 &41.6 &54.5 &66.5 &74.2 &64.1 &50.4 \\
\rowcolor{gray!20} \cellcolor{white} & FOEM & 8.35 &12.94 &74.9 &72.3 &41.4 &54.9 &67.8 &74.6 &64.3 &50.5 \\
\hline
\end{tabular}
\caption{Comparison of W4A4KV4 weight–activation quantization for various Llama models, based on SpinQuant.}
\label{tab:activation}
\end{table*}

\section{Experiment}
\label{sec:experiment}

We evaluated our FOEM against other advanced quantization approaches including GPTQ~\cite{frantar2022gptq} and GPTAQ~\cite{li2025gptaqefficientfinetuningfreequantization} on Llama 2~\cite{touvron2023Llama}, Llama3~\cite{grattafiori2024Llama}, Qwen3~\cite{yang2025qwen3}, Phi~\cite{li2023textbooks} and Mistral~\cite{jiang2023mistral7b} models. In addition, we also include the baseline results of RTN(round-to-nearest) for reference. For calibration, we randomly sampled 128 data sequences from the c4 dataset with a sequence length of 2048, which is a commonly used calibration set and standard sequence length in the quantization field. The quantization configuration included weight-only quantization with group size 128 and activation quantization with KV cache quantization, where the activation quantization used the pre-learned rotation matrix published by SpinQuant by default. The evaluation tasks include perplexity (PPL) on WikiText2~\cite{merity2016pointer} and C4~\cite{raffel2020exploring}, zero-shot accuracy on six established reasoning benchmarks (PIQA~\cite{bisk2020piqa}, Winogrande~\cite{sakaguchi2021winogrande}, ARC-Easy~\cite{clark2018think}, ARC-Challenge~\cite{clark2018think}, HellaSwag~\cite{zellers2019hellaswag}, and BoolQ~\cite{boolq}), and 5-shot MMLU~\cite{hendrycks2020measuring}. For the single hyperparameter $\beta$ in our method, we consistently set it to 0.1 as this value demonstrates robust performance across all model architectures and quantization configurations. For the coefficient $\alpha$ mentioned in the GPTAQ paper but not specified, we set it to 0.2 following the configuration in their official code repository. The quantization calibration process was conducted on a single NVIDIA A800-80GB GPU, while evaluating the 70B model required 2$\times$ A800 GPUs for testing. 
For each method and dataset, we used predefined hyperparameters and ran the experiments three times with different predetermined random seeds for data sampling, then recorded the average results.

\subsection{Accuracy Results}

\paragraph{Weight-Only Quantization.}
In light of GPTQ's exclusive emphasis on weight-only quantization, we first conducted a comprehensive evaluation within this paradigm. Our investigation spans both 3-bit weight quantization and 4-bit quantization, enabling thorough comparisons across two widely-adopted quantization scales. 

On the widely adopted Llama family of models, our approach achieves consistent improvements across all evaluation metrics, as shown in Table~\ref{tab:weight_only} and Table~\ref{tab:weight_only_1}. For instance, in the representative case of 3-bit quantization on Llama3-8B, FOEM reduces perplexity on WikiText2 and C4 by 0.60 and 0.45 respectively, compared to the state-of-the-art compensation-based method GPTAQ. Moreover, it boosts 0-shot accuracy on commonsense datasets to 65.5 percent, narrowing the gap to the full-precision model to approximately 5 percent. In the more practically deployable 4-bit setting, FOEM consistently matches or surpasses full-precision performance. For example, on Llama2-13B, FOEM achieves parity with the fp model on the MMLU benchmark.

As shown in Table~\ref{tab:weight_only_2}, FOEM also demonstrates strong performance on large language models beyond the Llama series. On Qwen3-8B~\cite{yang2025qwen3}, it reduces WikiText2 perplexity from 8.32 with GPTQ to 8.17. On Phi-1.5B~\cite{gu2024mamba}, FOEM is the only method capable of maintaining MMLU accuracy above 40.

\begin{table}[th!]
\centering
\setlength{\tabcolsep}{0.30mm}
\begin{tabular}{ll||c|cc|ccccccc|c}
\hline
\textbf{Model} & \textbf{Method} & \textbf{\#W} & \textbf{Wiki2$\downarrow$} & \textbf{c4$\downarrow$} & \textbf{0-shot Avg$\uparrow$} \\
\hline
\multirow{4}{*}{\centering Llama3-70B} 
& FP16 & 16 & 2.86 & 7.31 & 76.9 \\
\cdashline{2-6}
& GPTQ & 3 & 5.37 & 9.31 & 73.9 \\
& GPTAQ & 3 & 5.41 & 9.26 & 74.0 \\
\rowcolor{gray!20} \cellcolor{white} & FOEM & 3 & 5.36 & 9.21 & 74.1  \\
\hline
\end{tabular}
\caption{Weight-only quantization for Llama3-70B.}
\label{tab:weight_only_3}
\end{table}

Even for the large-scale Llama3-70B model, where performance remains relatively stable across different quantization methods, FOEM still achieves the best quantization results. It reduces C4 perplexity by 0.1 compared to GPTQ.

\paragraph{Weight-Activation Quantization.}
Our experimental analysis under W4A4KV4 configurations demonstrates the effectiveness of our method when combined with SpinQuant, the advanced rotation-based techniques for activation quantization, across multiple model scales. Notably, we directly utilize the publicly released pre-trained rotation matrices without any further tuning.
For the Llama series models, our approach consistently outperforms both GPTQ and GPTAQ on the MMLU benchmark and zero-shot evaluation tasks. For example, it improves the zero-shot average accuracy of Llama3-8B to 64.3. Although GPTQ and GPTAQ already achieve competitive performance that is close to the full-precision baseline in zero-shot settings, our method further narrows the accuracy gap and sets new state-of-the-art results.
Notably, the advantage of FOEM is even more pronounced in terms of perplexity. On W3A16 quantization for Llama3-8B, it reduces ppl by approximately 0.2 compared to other advanced compensation-based methods.

\begin{table}[th!]
\centering
\setlength{\tabcolsep}{2.28mm}
\begin{tabular}{l||c|cc|ccc}
\hline
\textbf{Method} & \textbf{$\beta$} & \textbf{Wiki2$\downarrow$} & \textbf{c4$\downarrow$} & \textbf{0-shot Avg$\uparrow$} \\
\hline

\multirow{1}{*}{\centering GPTQ} 
 & - & 9.86 & 12.94 & 64.4 \\
\hline
\multirow{10}{*}{\centering FOEM} 
 &\cellcolor{gray!20}0.1 &\cellcolor{gray!20}8.32 &\cellcolor{gray!20}12.37 &\cellcolor{gray!20}65.5 \\
 & 0.2 & 8.87 & 12.90 & 65.7 \\
 & 0.3 & 8.43 & 12.88 & 65.8 \\
 & 0.4 & 8.59 & 12.60 & 64.5 \\
 & 0.5 & 8.52 & 12.51 & 64.9 \\
\cdashline{2-5}
 & 0.6 & 9.67 & 12.72 & 64.6 \\
 & 0.7 & 9.79 & 12.80 & 64.2 \\
 & 0.8 & 10.09 & 13.09 & 61.5 \\
 & 0.9 & 9.63 & 13.26 & 62.3 \\
 & 1.0 & 10.37 & 14.41 & 61.5 \\
\hline
\end{tabular}
\caption{Sensitivity Analysis of $\beta$ on Llama3-8B.}
\label{tab:beta}
\end{table}

\paragraph{Sensitivity Analysis.}
We conduct a comprehensive evaluation of the impact of different settings of $\beta$ on the final quantization results in Table~\ref{tab:beta}. When $\beta$ is less than 0.5, FOEM consistently delivers significant performance gains. Although we set $\beta$ to 0.1 throughout other experiments, tuning this parameter can further enhance the effectiveness of FOEM.
However, when $\beta$ exceeds 0.5, the model’s performance degrades substantially after quantization. This aligns with our analysis in the methodology section, where we noted that excessively large gradients can amplify approximation errors and lead to suboptimal outcomes. This finding is also consistent with observations in GPTAQ, where the use of a properly chosen stabilization coefficient is essential for ensuring consistently positive effects.

\paragraph{Performance on the State Space Model.}
Beyond standard Transformer architectures, we evaluated SSM models, exemplified by Mamba~\cite{gu2024mamba}. As shown in Table~\ref{tab:Mamba}, applying W3A16 quantization to Mamba-1.4B with a default beta of 0.1, FOEM substantially outperforms GPTAQ, demonstrating strong cross-architecture generalization.

\begin{table}[th!]
\centering
\setlength{\tabcolsep}{5.98mm}
\begin{tabular}{l||c|cc|ccc}
\hline
\textbf{Method} & \textbf{$\beta$} & \textbf{Wikitext2 PPL$\downarrow$} \\
\hline
GPTAQ & - & 14.10 \\
\hline
\multirow{3}{*}{\centering FOEM} 
 & \cellcolor{gray!20}{0.1} & \cellcolor{gray!20}{13.91} \\
 & 0.2 & 14.06 \\
\cdashline{2-3}
 & 0.3 & 14.25  \\
\hline
\end{tabular}
\caption{Performance and sensitivity analysis of W3A16 quantization on Mamba-1.4B.}
\label{tab:Mamba}
\end{table}

\subsection{Efficiency Analysis}

Compared to GPTQ, our method only introduces a simple weight difference operation that requires no matrix multiplication and adds virtually no computational overhead relative to other steps in the quantization process. As shown in Table~\ref{tab:Llama3_8b_efficiency}, for weight-activation quantization on Llama3-8B, FOEM achieves nearly identical quantization time as GPTQ. In contrast to GPTAQ, which incurs significant additional computation and time overhead, FOEM not only offers substantially lower latency but also delivers superior accuracy. As shown in Table~\ref{tab:Llama3_8b_deploy}, when deploying the LLaMA3-8B model under W4A16 using vLLM~\cite{kwon2023efficient}, we achieve over 30\% speedup from weight-only quantization, in addition to the memory savings from compression. These empirical results demonstrate the effectiveness of incorporating first-order information, both in improving numerical precision and maintaining practical deployment efficiency.

\begin{table}[th!]
\centering
    \setlength{\tabcolsep}{1.8mm}
\begin{tabular}{ll||rcccc}
\hline
\textbf{Model} & \textbf{Method} & \textbf{Quant. Time (s)} & \textbf{Wiki2$\downarrow$}\\
\hline
\multirow{3}{*}{\centering Llama3-8B} 
& GPTQ & 825.50 & 8.55\\
& GPTAQ & 1112.20 & 8.50 \\
\rowcolor{gray!20} \cellcolor{white} & FOEM & 828.90 & 8.35 \\
\hline
\end{tabular}
\caption{Performance and Quantization Time for Llama3-8B.}
\label{tab:Llama3_8b_efficiency}
\end{table}

\begin{table}[th!]
\centering
    \setlength{\tabcolsep}{3.45mm}
\begin{tabular}{l||ccccc}
\hline
\textbf{Method} & \textbf{Input Tokens/s} & \textbf{Output Tokens/s}\\
\hline
FP & 184.11 & 470.11\\
\rowcolor{gray!20} FOEM & 250.26 & 616.01 \\
\hline
\end{tabular}
\caption{Inference speed for Llama3-8B on W4A16.}
\label{tab:Llama3_8b_deploy}
\end{table}

\section{Conclusion.}

In this paper, we introduce FOEM, a novel PTQ method that incorporates first-order terms in the Taylor expansion of quantization loss to enable more accurate error compensation. Although full-precision models are typically assumed to be well-optimized, we observe that the application of compensation terms during quantization causes the remaining unquantized weights to deviate from their original values. As a result, the latent weights can exhibit non-negligible gradients even before quantization. To address this issue, FOEM integrates the first-order term into the Lagrangian formulation for joint optimization. Based on the derived theoretical expression, we approximate the gradient term efficiently using the difference between the original full-precision weights and the current latent weights, significantly reducing computational cost and eliminating the need for calibration data. Additionally, we utilize precomputed Cholesky factors to reconstruct the inverse Hessian submatrices on the fly, ensuring computational efficiency.
FOEM consistently outperforms GPTQ across a wide range of benchmark evaluations. For example, in 3-bit weight-only quantization, FOEM reduces the perplexity of Llama3-8B by 17.3\%. Furthermore, FOEM is compatible with state-of-the-art quantization strategies such as SpinQuant, offering enhanced performance while maintaining efficiency. This makes FOEM a promising solution for the practical and accurate deployment of large language models, demonstrating its broad application potential.

\section{Acknowledgments}
This work was supported by the National Natural Science Foundation of China (Nos. 62476018, 62306025, 92367204), the Fundamental Research Funds for the Central Universities, the Beijing Municipal Science and Technology Project (No. Z231100010323002), and Swiss National Science Foundation (SNSF) project 200021E\_219943 Neuromorphic Attention Models for Event Data (NAMED).

\bibliography{aaai2026}

\end{document}